\documentclass{article}

\usepackage{microtype}
\usepackage{graphicx}
\usepackage{subfigure}
\usepackage{booktabs} 

\usepackage{hyperref}
\usepackage{bm}

\usepackage[accepted]{icml2024}

\usepackage{amsmath}
\usepackage{amssymb}
\usepackage{mathtools}
\usepackage{amsthm}
\usepackage{multirow}

\usepackage{afterpage}

\usepackage[capitalize,noabbrev]{cleveref}

\theoremstyle{plain}

\theoremstyle{definition}

\theoremstyle{remark}

\usepackage[textsize=tiny]{todonotes}
\usepackage{pifont}
\icmltitlerunning{In-Context Decision Transformer: Reinforcement Learning via Hierarchical Chain-of-Thought}

\begin{document}

\twocolumn[
\icmltitle{In-Context Decision Transformer: Reinforcement Learning via Hierarchical Chain-of-Thought}



\icmlsetsymbol{equal}{*}

\begin{icmlauthorlist}
\icmlauthor{Sili Huang}{sch1,sch2}
\icmlauthor{Jifeng Hu}{sch1}
\icmlauthor{Hechang Chen}{equal,sch1}
\icmlauthor{Lichao Sun}{sch3}
\icmlauthor{Bo Yang}{equal,sch2}
\end{icmlauthorlist}

\icmlaffiliation{sch1}{School of Artificial Intelligence, Jilin University, China}
\icmlaffiliation{sch2}{Key Laboratory of Symbolic Computation and Knowledge Engineering of Ministry of Education, Jilin University, China}
\icmlaffiliation{sch3}{Lehigh University, Bethlehem, Pennsylvania, USA}

\icmlcorrespondingauthor{Bo Yang}{ybo@jlu.edu.cn}
\icmlcorrespondingauthor{Hechang Chen}{chenhc@jlu.edu.cn}

\icmlkeywords{Machine Learning, ICML}

\vskip 0.3in
]



\printAffiliationsAndNotice{\icmlEqualContribution} 

\begin{abstract}
In-context learning is a promising approach for offline reinforcement learning (RL) to handle online tasks, which can be achieved by providing task prompts. Recent works demonstrated that in-context RL could emerge with self-improvement in a trial-and-error manner when treating RL tasks as an across-episodic sequential prediction problem. Despite the self-improvement not requiring gradient updates, current works still suffer from high computational costs when the across-episodic sequence increases with task horizons. To this end, we propose an In-context Decision Transformer (IDT) to achieve self-improvement in a high-level trial-and-error manner. Specifically, IDT is inspired by the efficient hierarchical structure of human decision-making and thus reconstructs the sequence to consist of high-level decisions instead of low-level actions that interact with environments. As one high-level decision can guide multi-step low-level actions, IDT naturally avoids excessively long sequences and solves online tasks more efficiently. Experimental results show that IDT achieves state-of-the-art in long-horizon tasks over current in-context RL methods. In particular, the online evaluation time of our IDT is \textbf{36$\bm\times$} times faster than baselines in the D4RL benchmark and \textbf{27$\bm\times$} times faster in the Grid World benchmark.
\end{abstract}


\section{Introduction}

Large transformer models \cite{4} have shown impressive abilities across a variety of domains, including text \cite{5}, image \cite{6}, and audio \cite{7}. 
In the field of reinforcement learning (RL), large transformer models can treat the RL tasks as a type of sequential prediction problem, which has proven successful in using solely offline training \cite{14,27}.
A notable shortcoming lies with these methods to self-improve when employed in online environments. To overcome this, in-context RL methods have been introduced, which enable continued policy improvement \cite{8}.

Recent works demonstrated that in-context RL can automatically improve its performance in a trial-and-error manner when across-episodic contexts serve as prompt conditions \cite{3}. The construction of the across-episodic context is flexible and easy to implement, such as a chain of experience that consists of multiple historical trajectories arranged in ascending order of returns \cite{1}. Despite the progress made, current methods are mostly limited to short-horizon tasks with less than 100 timesteps \cite{8}. This arises from (1) the quadratic complexity of the self-attention mechanism and (2) the significant increase in the length of sequences caused by across-episodic contexts. Such huge computational costs severely limit in-context RL to apply the trial-and-error ability on traditional RL tasks, which often reach 1000 timesteps, such as MuJoCo \cite{13} and Atari \cite{15}.

In fact, trial-and-error is the central idea of modern RL algorithms \cite{16}. It is an animal behavior originated by a psychologist \citet{17} who considers trial-and-error as an elementary way of combining search and memory. Correspondingly, the across-episodic contexts provide memory, and the self-attention mechanism reviews historical actions in the memory to search for better actions. However, human decision-making is more complex and operates on multiple levels of temporal abstraction \cite{12}. For example, travelers tend to decide on their budget first, then their mode of transportation, right down to the smallest action. Inspired by this idea, a natural perspective emerges: 
\vspace{-5pt}
\begin{center}
\emph{``Can human multi-level decision-making bring out a more efficient trial-and-error?"}
\end{center}
\vspace{-5pt}


\begin{figure}
    \centering
    \includegraphics[width=1.0\columnwidth]{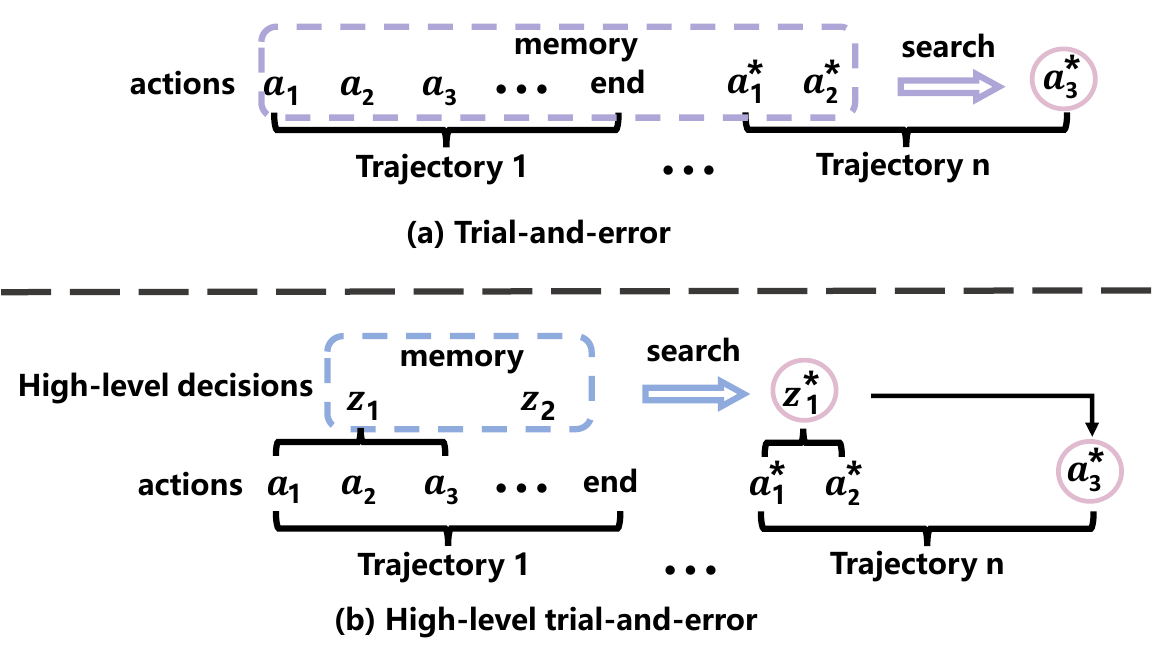}
    \caption{Trial-and-error comparison of minimal actions and high-level decisions, where * denotes better results. (a) In the trial-and-error process, the memory consists of the smallest actions from experiences and serves as context to search for better action. (b) In the high-level trial-and-error process, the memory and search act on high-level decisions. Since one high-level decision controls multiple actions, we can use smaller memory to preserve experiences and search for better decisions with less computational costs.}
    \label{motivation}
\end{figure}

As one high-level decision can guide multi-step low-level actions, it can considerably shorten across-episodic contexts and, therefore, significantly alleviate the computational costs. In this view, we aim to explore a high-level trial-and-error process, as shown in Figure~\ref{motivation}. 
However, since the model generates high-level decisions rather than low-level actions that interact with the environment, our first challenge is ensuring that better high-level decisions can encourage better low-level actions.
In addition, the in-context RL is trained with supervised losses, which predicts each next step conditioned on the past steps in the sequence.
Unlike low-level actions, the high-level decision is an abstract concept that is usually not directly observable from training data.

To this end, we propose an efficient in-context RL method called In-context Decision Transformer (IDT). 
Specifically, IDT consists of Making Decisions, Decisions to Go, and Reviewing Decisions modules to mimic the high-level trial-and-error process. First, the Making Decisions module is a decoder-only transformer that generates high-level decisions autoregressively, where the high-level decision is represented by a vector sampled from a multivariate Gaussian distribution. Then, the generated high-level decisions are fed to the Decisions to Go module, which is also a decoder-only transformer to generate low-level actions autoregressively. 
The output of the Making Decisions module serves as a conditional input to the Decisions to Go module, ensuring high-level decisions correctly guide low-level actions.
To fill in the missing high-level decisions in the training data, we designed the Reviewing Decisions module to encode high-level decisions from sequences of low-level actions.
All three modules are learned end-to-end by predicting the low-level actions from the training data.

Our contributions are as follows: (1) We propose IDT, an in-context RL method that emerges with high-level trial-and-error ability. IDT can learn by directly combining sub-optimal data and efficiently improving itself through multiple trials at test time. (2) Compared to the contexts consisting of the smallest actions, IDT significantly shortens the evaluation time by \textbf{36$\bm\times$} times on the D4RL baselines \cite{13} and \textbf{27$\bm\times$} times on the Large Grid World environments \cite{14}. 
(3) IDT can achieve state-of-the-art results with less training costs, especially outstanding in long-horizon tasks. 

\section{Related Work}

\noindent\textbf{Transformer for Decision-Making.}~~
In general, reinforcement learning was proposed as a fundamentally online paradigm \cite{38}. The nature of online learning comes with some limitations when meeting the applications for which it is impossible to gather online data and learn simultaneously, such as autonomous driving. To this end, offline RL proposes that the agent can learn from a fixed dataset of previously collected data without gathering new data during learning \cite{39,40,41,42}. In the context of offline RL, recent works explored using transformer-based policy by treating RL tasks as a type of sequential prediction problem. Among them, a decision transformer \cite{2} is proposed to model trajectories as sequences and autoregressively predicts action conditioning on desired return-to-go, past states, and actions. Trajectory transformer \cite{43} demonstrated that transformer could learn single-task policies from offline data. Subsequently, the multi-game decision transformer \cite{14} and Gato \cite{27} further showed that transformer-based policies could address multi-tasks in the same domain and cross-domain tasks. However, these works focused on distilling expert policies from offline data and failed to enable self-improvement like IDT. Instead, when the offline data are sub-optimal, or the agent is required to adapt to new tasks, the multi-game decision transformers need to finetune the model parameters while Gato is required to get prompted with expert demonstrations.

\noindent\textbf{Meta RL.}~~
IDT falls into the category of methods of learning to learn, which is also known as meta-learning. More precisely, recent in-context RL methods can be categorized as in-context meta-RL methods. The general idea of learning self-improvement has a long history in RL but is limited to hyper-parameters in the early stages \cite{31}. In-context meta-RL methods \cite{33,32} are commonly trained in the online setting by maximizing multi-episodic value functions with memory-based architectures through environment interactions. Another online meta-RL attempts to find good network parameter initializations and then quickly adapt through additional gradient updates \cite{46,45}. More recently, meta-RL has seen substantial breakthroughs, from performance gains on popular benchmarks to offline settings, such as Bayesian RL \cite{36} and optimization-based meta-RL \cite{37}. Considering the difficulty of a completely offline setting, recent work has explored hybrid offline-online settings \cite{34,35}. IDT is similar to the hybrid offline-online setting, but the online phase does not involve gradient updates.

\noindent\textbf{In-Context RL.}~~
In-context RL is the one that addresses tasks by providing prompts or demonstrations \cite{2,43}. By training agents at a large scale, transformer-based policies usually have the ability to learn in context \cite{14,27}. The learning process is performed entirely in context and does not involve parameter updates of neural networks. In this work, we consider incremental in-context RL that involves learning from one's own behaviors through a trial-and-error manner. \citet{8} proposed Algorithm Distillation (AD) that automatically improves its performance in a trial-and-error manner by providing multiple historical trajectories. Subsequently, \citet{44} proposed a Decision-Pretrained Transformer, which trains the agent to find optimal behaviors faster by only predicting the optimal trajectory. More recently, \citet{1} further demonstrated that across-episodic contexts encourage large transformer models' emerging trial-and-error behaviors. However, these methods focus on the smallest action level, which causes across-episodic contexts to induce too-long sequences and suffer from huge computational costs. In contrast, IDT explores the trial-and-error ability of high-level decisions, which can significantly shorten the length of across-episodic contexts and, therefore, alleviate the computational costs arising from the self-attention mechanism.

\section{Preliminaries}

\noindent\textbf{Partially Observable Markov Decision Process.}~~
We consider learning problems in the context of Partially Observable Markov Decision Processes (POMDP) represented by the tuple $\mathcal{M}=(\mathcal{S,O,A},P, \mathcal{R})$. The POMDP tuple consist of states $s\in \mathcal{S} $, observations $o\in \mathcal{O} $, actions $a\in \mathcal{A} $, rewards $r\in \mathcal{R} $, and transition probability function $P (s_{t+1}|s_t,a_t)$, where $t$ is an integer denoting the timestep. In environments described by a POMDP, at each timestep $t$ the agent receives the observation $o_t$, selects an action $a_t \sim \pi(\cdot|o_t)$ from its policy, and then receives the next observation $o_{t+1}$. A trajectory is a sequence of observations, actions, and rewards and is denoted by $\tau=(o_0,a_0,r_0,\dots,o_T,a_T,r_T)$. The return of a trajectory at timestep $t$, $R_t = \sum_{t'=t}^Tr_t'$, is calculated as the sum of future rewards from that timestep.
In addition, a completion token $d_t$, a binary identifier, is used to indicate whether a trajectory ends at time $t$.

\noindent\textbf{Hierarchical Reinforcement Learning.}~~
RL algorithms aim to maximize the expected return $\mathrm{E}[\sum_{t=0}^Tr_t]$ throughout an agent's lifetime or training episodes. In long-horizon tasks, standard RL methods suffer from poor performance due to the exponentially growing exploration space. Hierarchical RL decomposes the long-horizon task into subproblems or subtasks such that a high-level policy learns to perform the task by choosing optimal subtasks as the high-level decisions \cite{18}. High-level decisions can be designed as discrete or continuous forms. The discrete form can select multiple independent low-level policy models \cite{19}, while the continuous form usually serves as additional conditions to control a general low-level policy model \cite{10}. Since the transformer-based policy is a conditional generative model, it is naturally adapted to high-level decisions in the continuous form, such as the return-to-go condition in the decision transformer \cite{2}. In this work, we use a vector $\textbf{z}$ to represent high-level decisions and assume that it is sampled from a multivariate Gaussian distribution.

\noindent\textbf{Transformers.}~~
The Transformer \cite{4} architecture consists of multiple layers of self-attention operation and MLP. The self-attention begins by projecting input data $X$ with three separate matrices onto $D$-dimensional vectors called queries $Q$, keys $K$, and values $V$. These vectors are then passed through the attention function:
\begin{equation}
    \mathrm{Attention}(Q,K,V)=\mathrm{softmax}(QK^T/\sqrt{D})V.
\label{self-attention}
\end{equation}
The $QK^T$ term computes an inner product between two projections of the input data $X$. The inner product is then normalized and projected back to a $D$-dimensional vector with the scaling term $V$. Transformers utilize self-attention as a core part of the architecture to process sequential data \cite{22,21}. In this work, we use GPT \cite{23} architecture that modifies the transformer with a causal self-attention mask to focus on the previous tokens in the sequence ($j\in[1,i]$), enabling us to do autoregressive generation at test time.
\begin{figure*}
    \centering
    \includegraphics[width=1.7\columnwidth]{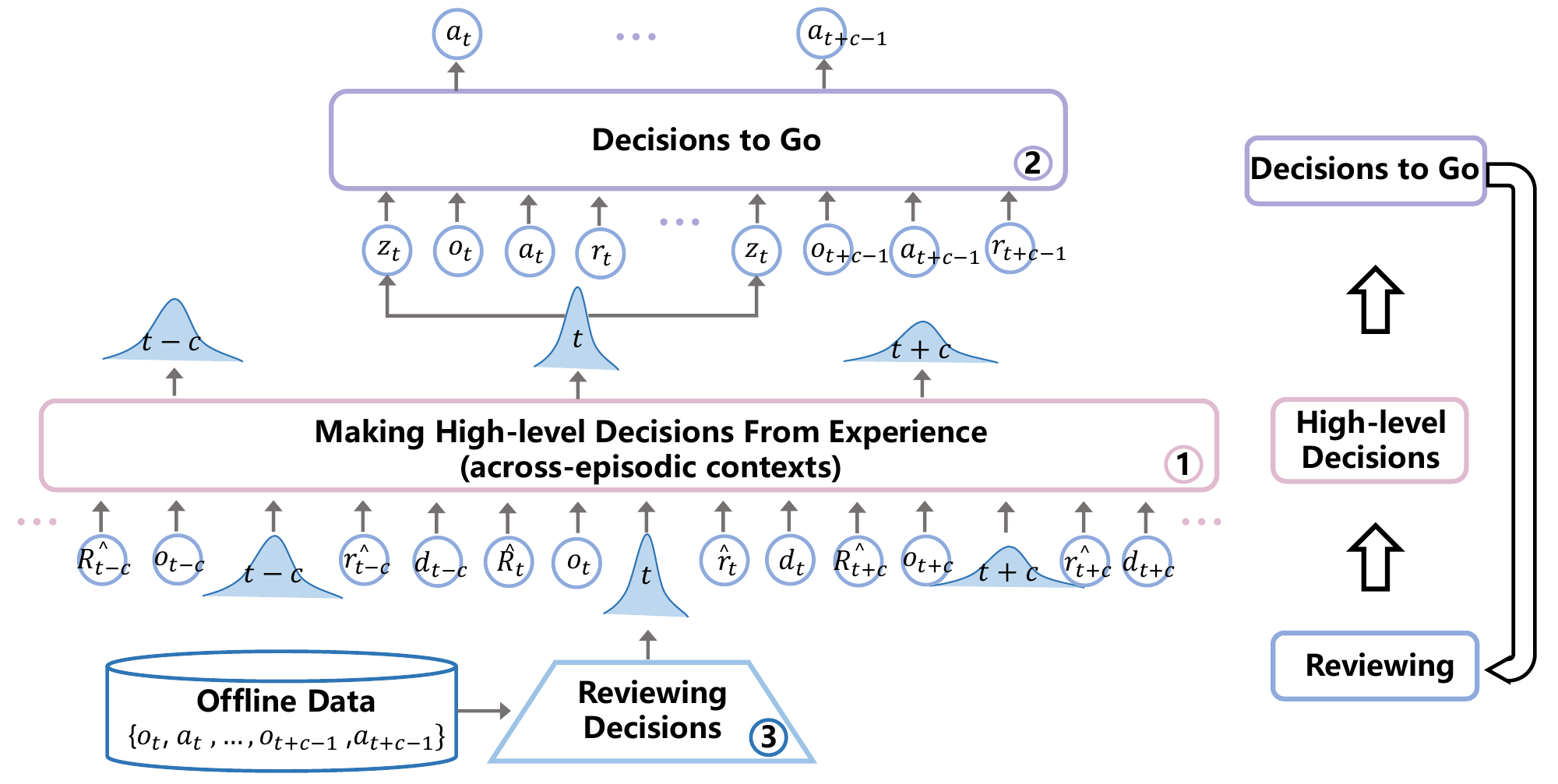}
    \caption{The architecture of IDT is designed into three modules to simulate the high-level trial-and-error process. First, the (1) Making Decisions module predicts a high-level decision by providing across-episodic contexts, where across-episodic contexts contain multiple trajectories arranged in ascending order of the total rewards. Then, the (2) Decisions to Go module predicts actions for $c$ steps conditioned on the predicted high-level decision. Finally, the (3) Reviewing Decisions module reviews the executed actions to serve as an experience for the next cycle. Note that the Reviewing Decisions encodes the true label of high-level decisions from offline data at training while encodes from the executed actions at testing.
    }
    \label{IDT}
\end{figure*}

\section{Method}
In this section, we present IDT, which models a high-level trial-and-error process through a hierarchical chain of experience, as summarized in Figure ~\ref{IDT}.

\subsection{Chain of Experience}
The key factors that influence our modeling on how to represent trajectories are (1) the ability of transformers to uncover meaningful patterns from multiple trajectories and (2) the capacity to improve itself conditioned on experience. The basic elements of trajectories are observations $o$, actions $a$, rewards $r$, and completion token $d$. As modeling rewards is a nontrivial task, we aim to have the model generate actions based on the target returns $\hat{R}_0$ \cite{2}, which can be updated using rewards $\hat{R}_t=\hat{R}_0-\sum_{j=0}^tr_j$. Therefore, the following trajectory representation is amenable to autoregressive training and generation:
\begin{equation}
    \tau = (\hat{R}_0,o_0,a_0,r_0,d_0,\dots,\hat{R}_T,o_T,a_T,r_T,d_T).
\end{equation}

To facilitate the model to achieve target return, we construct across-episodic contexts that consist of multiple trajectories for self-improvement during test time \cite{1}. This idea arises from the approach called chain of hindsight \cite{24}, which trains language models from human feedback by conditioning on positive indicators and negative-rated examples to predict corresponding positive-rated examples. In the RL tasks, the positive indicator is the target return, and previous trajectories serve as negative-rated examples to predict the trajectory with higher returns.

Specifically, the chain of experience is represented by $n$ trajectories: $s=(\tau^{1},\tau^{2},\dots,\tau^{n})$ where
\begin{equation}
    \tau^{i}= (\hat{R}_0^i,o_0^i,a_0^i,r_0^i,d_0^i,\dots,\hat{R}_T^i,o_T^i,a_T^i,r_T^i,d_T^i).
\label{coe}
\end{equation}

The trajectories are ascending sorted according to their total rewards, i.e., $\sum_{t=0}^T{r_t^1}\leq\sum_{t=0}^T{r_t^2}\leq\dots\leq\sum_{t=0}^T{r_t^n}$. For all $n$ trajectories, the initial target return $\hat{R}_0^i$ equals the max total reward, i.e., the last trajectory $\hat{R}_0^n=\sum_{t=0}^T{r_t^n}$.

\subsection{Hierarchical Chain of Experience}
After building across-episodic contexts based on the chain of experience, in-context RL can automatically improve its performance at evaluation time by rolling trajectories in a trial-and-error manner. However, this suffers substantial computational costs when the horizon of tasks increases.

Since total rewards are obtained at the end of episodes, it is more difficult to evaluate and improve a policy model in long-horizon tasks. In traditional RL methods, an effective solution is to decompose complex tasks into several sub-problems by incorporating hierarchical structures  \cite{10}. The high-level policy only needs to generate a signal once to control the low-level policy to generate multi-step actions. This allows (1) the high-level policy to receive feedback faster, as if working on a short-horizon task, and (2) the low-level policy only needs to consider how to better implement the sub-tasks generated by the high-level decision. Although the hierarchical structure can be optimized end-to-end by reward signals, the trial-and-error process in in-context RL is more complicated.

As psychologist Edward Thorndike mentioned, the trial-and-error process includes two parts \cite{16}, memory and search. The high-level decision plays an important role that is closely connected with memory and search. A high-level decision is generated from the search process and directly affects the quality of low-level executed actions. Subsequently, it also serves as the memory for future searches. Therefore, we designed three modules to realize a high-level trial-and-error process: Making Decisions, Decisions to Go, and Reviewing Decisions.

\noindent\textbf{Making Decisions.}~~
The purpose of the Making Decisions module is to generate high-level decisions autoregressively, where the high-level decision is represented by a vector $\textbf{z}$ sampled from a multivariate Gaussian distribution. As the quality of $\textbf{z}$ directly relates to low-level actions $a$, a better high-level decision $\textbf{z}$ is critical for inducing better low-level actions $a$. Therefore, we reconstruct across-episodic contexts represented as a high-level chain of experience $s_h=(\tau_h^{1},\tau_h^{2},\dots,\tau_h^{n})$. Each $\tau_h^{i}$ is denoted as:
\begin{equation}
\begin{aligned}
\tau_h^{i}=(&\hat{R}_0^i,o_0^i,\textbf{z}_0^i,\hat{r}_0^i,d_0^i,\hat{R}_{c}^i,o_{c}^i,\textbf{z}_{c}^i,\hat{r}_{c}^i,d_{c}^i,\\
&\dots,\hat{R}_{kc}^i,o_{kc}^i,\textbf{z}_{kc}^i,\hat{r}_{kc}^i,d_{kc}^i),
\label{hcoe}
\end{aligned}
\end{equation}
where each high-level decision $\textbf{z}$ is generated every $c$ steps, $T-c\leq kc \leq T$, and $\hat{r}_c^i=\sum_{t=c}^{2c-1}{r_t^i}$ is the sum of $c$ steps rewards. By comparing with Equation \eqref{coe}, the high-level chain of experience can considerably shorten the length of contexts and, therefore, significantly alleviate the computational complexity of the self-attention mechanism.

\noindent\textbf{Decisions to Go.}~~
Based on high-level decisions, the Decisions to Go module is designed to generate low-level actions that can interact with environments. Since the transformer-based policy is a conditional generative model, we can build a low-level context that contains high-level decisions to control the low-level actions. The low-level context is represented as:
\begin{equation}
\begin{aligned}
\tau_l^{i,j}=&(\textbf{z}_j^i,o_j^i,a_j^i,r_j^i, \textbf{z}_{j}^i,o_{j+1}^i,a_{j+1}^i,r_{j+1}^i, \\ &\dots,\textbf{z}_{j}^i,o_{j+c-1}^i,a_{j+c-1}^i,r_{j+c-1}^i),
\label{lcoe}
\end{aligned}
\end{equation}
where each $\tau_l^{i,j}$ starts from the generation step $j\in\{0,c,\dots,kc\}$ of high-level decisions and completes $c$ steps low-level actions in the trajectory $\tau^i$. In particular, we introduce the reparameterization trick \cite{20} into high-level decisions to ensure backpropagation through the Decisions to Go module to the Making Decisions module. 

\noindent\textbf{Reviewing Decisions.}~~
The autoregressive training of the conditional generation model is achieved by predicting each token in the sequence. 
For example, when the transformer model is trained to generate $a_t$, we need to provide the action label at time $t$ and condition it on the historical actions $\{a_0,\dots,a_{t-1}\}$. However, the supervisory signals about high-level decisions $\textbf{z}$ are not directly observable from the sequence, as shown in Equation~\eqref{hcoe}. For instance, when the transformer model is trained to generate $\textbf{z}_{j}$ ($j\in\{0,c,\dots,kc\}$), we have neither the true label of $\textbf{z}_{j}$ nor the previous high-level decisions $\{\textbf{z}_0,\textbf{z}_c,\dots,\textbf{z}_{j-c}\}$. 

To this end, we replace the true label of $\textbf{z}_{j}$ with the gradients from the Decisions to Go module trained to generate the following $c$ steps of actions $\{a_{j},a_{j+1},\dots,a_{j+c-1}\}$. For the previous high-level decisions $\{\textbf{z}_0,\textbf{z}_c,\dots,\textbf{z}_{j-c}\}$, we introduce the Reviewing Decisions module to encode the label from low-level actions. 
As the high-level decisions induce low-level actions, the low-level actions should be able to infer high-level decisions inversely. Specifically, to infer a previous high-level decision $\textbf{z}_t \in \{\textbf{z}_0,\textbf{z}_c,\dots,\textbf{z}_{j-c}\}$, we first utilize the self-attention operation to aggregate the information of $a_{t+c-1}$ and $\{o_t,a_t,\dots,o_{t+c-1},a_{t+c-1}\}$. Then, we apply a linear layer to encode $\textbf{z}_t$ from the aggregated information. Note that the Reviewing Decision module is not required to perform autoregressive generation, so any sequence model, such as LSTM, can replace it.

By combining the above three modules, IDT can automatically improve its performance at evaluation time by rolling trajectories in a high-level trial-and-error manner. We now introduce the implementation details of IDT, including architecture, training, and testing.

\begin{figure*}
    \centering
    \includegraphics[width=1.9\columnwidth]{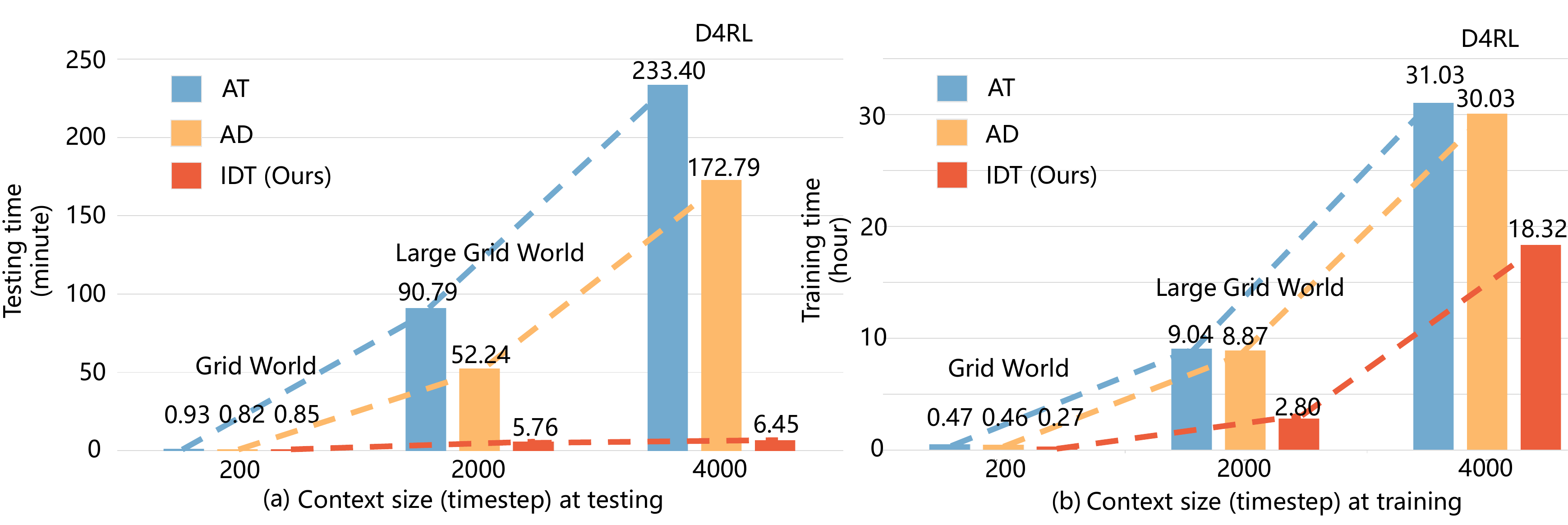}
    \caption{Results for (a) testing and (b) training times. We report the training time per 10k gradient updates, the testing time for 50 episodes over Grid World, and 10 episodes over D4RL. Note that we use the number of steps to measure the context size here. The number of tokens per step may vary depending on the algorithm. Each step in AD contains 4 tokens: observation, action, reward, and completion. IDT's Making Decisions module and AT have an extra return-to-go token. As the task length increases, the context length is forced to grow exponentially, resulting in a square increase in computational costs. In contrast, IDT reconstructs the sequence to consist of high-level decisions. Therefore, the context is smaller than one episode length, significantly reducing computational costs.}
    \label{cost_figure}
\end{figure*}

\subsection{Implementation of IDT}

\noindent\textbf{Architecture.}~~
We feed $n$ trajectories into the Making Decisions module, which results in $5\times n\times T/c$ tokens, with one token for each of the five modalities: desired target return, observation, high-level decision, reward, and completion. In the Decisions to Go module, we feed $4\times c$ tokens, with one token for each of the four modalities: high-level decision, observation, action, and rewards. In the Reviewing Decisions module, we feed $2\times c$ tokens, with one token for each of the two modalities: observation and action. To create the token embeddings, we train a linear layer for each modality, which transforms the raw inputs into the desired embedding dimension, followed by layer normalization \cite{26}. Finally, the tokens are processed by a GPT model that predicts future high- and low-level action tokens through autoregressive modeling.

\noindent\textbf{Training and Testing.}~~
During training, we are given a dataset of offline trajectories, where the trajectories can be suboptimal.
In each iteration, we sample minibatches of trajectories from the dataset. 
The Reviewing Decisions module first encodes each true high-level decision $\textbf{z}$ from the minibatch every $c$ steps.
Then, the Making Decisions module predicts the high-level decision $\textbf{z}_t$ given the input token $o_t$ and past trajectories. Finally, the Decisions to Go module autoregressively predicts $c$ steps of low-level actions $\{a_t,\dots,a_{t+c-1}\}$ given $\textbf{z}_t$ and $\{o_t,\dots,o_{t+c-1}\}$. The low-level actions are evaluated with either cross-entropy loss or mean-squared error, depending on whether the actions are discrete or continuous. The losses from each timestep are averaged and updated in all three modules end-to-end. At test time, we roll out the IDT with multiple trajectories and report the largest return among trajectories. Following the configuration from related works \citet{1,8}, we set a context size across $n=4$ episodes.
Note that the task horizons $T$ used in this work range from 20 steps to 1000 steps, and the maximum context size reaches 20000 tokens.
The pseudocode for IDT is summarized in Appendix~\ref{appendix A}. Source code and more hyperparameters are described in Appendix~\ref{appendix B}.

\section{Experiments}
\noindent\textbf{Dataset: Grid World.}~~
In this section, we first consider the discrete control environments from the Grid World \cite{14}, which is a commonly used benchmark for recent in-context RL methods. The environments support many tasks that cannot be solved through zero-shot generalization after pre-training because these tasks cannot be inferred easily from the observation. The episode of each task is short enough to train a transformer-based policy with across-episodic contexts feasibly. Specifically, we consider the four evaluation environments: Darkroom, Darkroom Hard, Darkroom Dynamic, and Dark Key-to-Door.

The evaluation environments provide a 2D discrete POMDP where an agent spawns in a room and must find a goal location. The agent only observes its own $(x,y)$ coordinates but does not know the goal location, which is required to deduce it from the rewards received. The room dimensions are $9\times9$ with the agent's possible actions, including moving one step either left, right, up, down, or staying idle. In Darkroom, an episode lasts 20 steps, and the agent can obtain a reward ($r=1$) each time the goal is achieved. The Darkroom Hard and Darkroom Dynamic are two variants of Darkroom. In the Darkroom Hard, agents only obtain a reward when the goal is achieved first. In the Darkroom Dynamic, the goal is fixed to a corner, but the action space is randomly permuted. In the Dark Key-to-Door, the length of an episode is 50, where the agent is required to locate an invisible key to receive a one-time reward first and then identify an invisible door to obtain another one-time reward.

In addition, we create a variant of Large Darkroom, Large Darkroom Hard, Large Darkroom Dynamic, and Large Darkroom Key-to-Door, where the coordinate space of each environment is expanded to $40\times40$, and the episode length is expanded 10 times. The dataset is collected from learning histories that are generated by training gradient-based RL algorithms, such as Deep Q-Network \cite{28}. For each environment, we randomly create 60 tasks from the coordinate space and collect data for 1 million timesteps. 

\begin{figure*}
    \centering
    \includegraphics[width=1.8\columnwidth]{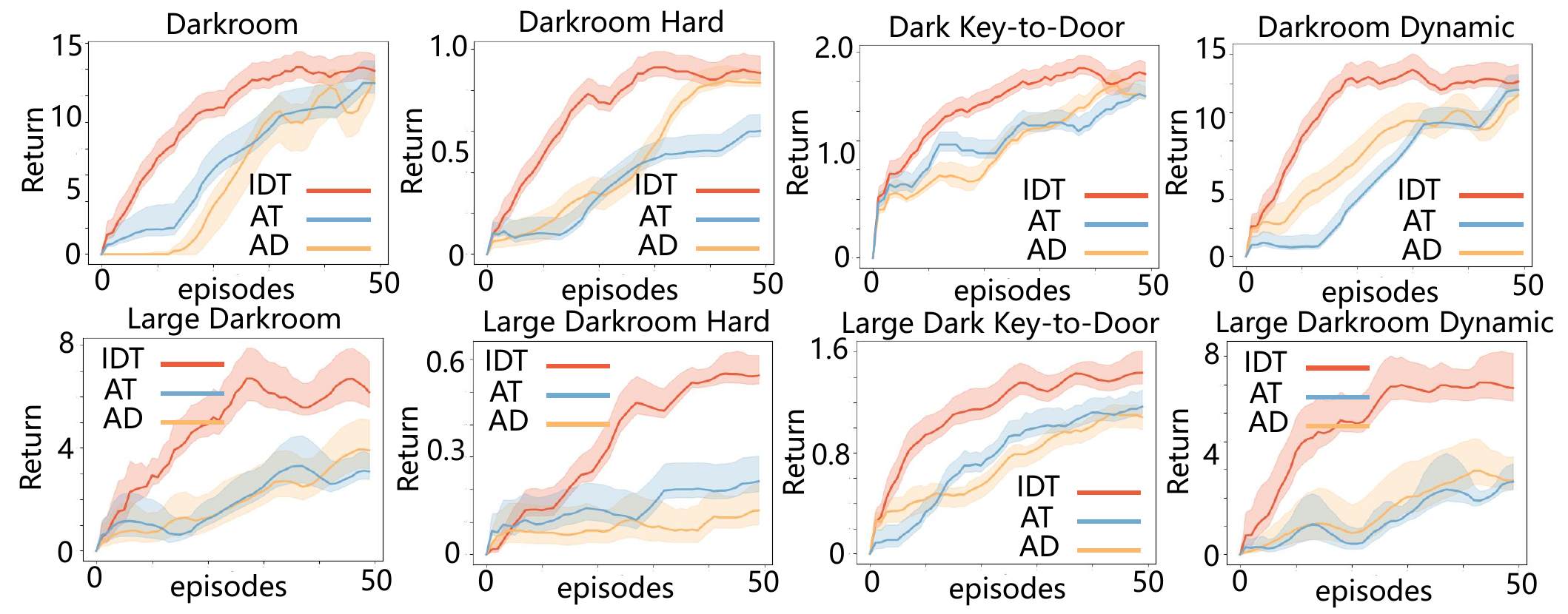}
    \caption{Results for Grid World. An agent is expected to solve a new task by interacting with the environments for 50 episodes without online model updates. Based on high-level decisions, our method outperforms both AT and AD, which rely on across-episodic contexts with the smallest actions. In particular, IDT has significant advantages in handling long-horizon tasks.}
    \label{grid world}
\end{figure*}

\noindent\textbf{Dateset: D4RL.}~~
D4RL \cite{13} is a commonly used offline RL benchmark, including continuous control tasks. The different dataset settings are described below.

\begin{itemize}
    \item Medium: 1 million timesteps generated by a ``medium" policy that performs approximately one-third as well as an expert policy.
    \item Medium-Replay: 1 million timesteps collected from the replay buffer of an agent trained to the performance of a ``medium" policy.
    \item Medium-Expert: It consists of 1 million timesteps generated by the ``medium" policy and another 1 million timesteps generated by the expert policy.
\end{itemize}

The dataset is collected from Mujoco environments, including HalfCheetah, Hopper, and Walker. The episode length in D4RL is 1000, which is far more than that of Grid World. Therefore, current in-context RL methods require huge computational costs in D4RL, even though it is a commonly used baseline for conventional RL algorithms.

\noindent\textbf{Baselines.}~~
In this section, we investigate the performance and efficiency of IDT relative to in-context RL, dedicated offline RL, and imitation learning algorithms. Our baselines can be categorized as follows:

\begin{itemize}
    \item In-context RL: These methods use the transformer to model trajectory sequences and predict actions autoregressively. We compare with recent methods, Agentic Transformer (AT) \cite{1} and Algorithm Distillation (AD) \cite{8}, which proposed across-episodic contexts with the smallest actions.
    \vspace{-2pt}
    \item Temporal-difference learning: Most temporal-difference (TD) learning methods use an action space constraint or value pessimism and will serve as faithful comparisons to IDT, representing standard RL methods. We consider state-of-the-art TD3+BC \cite{29} that is demonstrated to be effective on D4RL.
    \vspace{-2pt}
    \item Imitation learning: Imitation learning methods similarly utilize supervised losses for training, such as Behavior Cloning (BC) \cite{30} and Decision Transformer (DT) \cite{2}. Following AT, we compare with BC-10$\%$, which is shown to be competitive with state-of-the-art on D4RL. DT also uses a transformer to predict actions autoregressively but is limited to a single episode context.
\end{itemize}

For all comparison methods, we adhere closely to the original hyper-parameter settings. To evaluate IDT and other in-context RL algorithms, we roll out 10 episodes in D4RL and 50 episodes in Grid World. For each result, we report mean and standard error across 5 random seeds.

\subsection{Evaluation of Computing Costs}
An important property of in-context RL is that it can improve itself without expensive gradient updates. However, the computational costs of forward propagation are hidden in short-horizon tasks. Therefore, we reported the training time per 10k gradient updates, the evaluation time for 50 episodes over Grid World, and 10 episodes over D4RL. As shown in Figure~\ref{cost_figure}, our IDT has efficient training and significantly reduces the testing time compared to the baselines, approximately \textbf{36$\bm\times$} times faster in D4RL and \textbf{27$\bm\times$} times faster in large Grid World. More detailed results for each task are described in Appendix~\ref{additional experiments}.

As the task length increases, the evaluation time of AT and AD grows quadratically.
This is because the across-episodic contexts multiply the sequence length, leading to intolerable computational costs in the self-attention mechanism. The AT algorithm requires four episodes for trial-and-error, where each episode reaches 1000 steps in D4RL, and each step contains 5 tokens. 
Therefore, each step of AT generation requires scanning 20k tokens.
Since the AD algorithm reduces a return-to-go token at each step, the training and testing time are both less than AT.
In contrast, IDT reconstructs the sequence to consist of high-level decisions, and thus, the context is smaller than one episode length. As a result, IDT is significantly lower than baselines at both training and testing times.

\begin{table*}
\caption{Results for D4RL datasets. IDT outperforms both in-context RL (AT and AD) and supervised learning (BC) and performs competitively with conventional RL algorithms (TD3+BC and TD3) on almost all tasks.
}
\begin{center}
\resizebox{0.95\textwidth}{!}{
\begin{tabular}{lr|rrrrrrr}
\toprule
\specialrule{0em}{1.0pt}{1.0pt}
\toprule
Dataset & Environment & BC-10$\%$ & TD3+BC & TD3 & DT & AT & AD& Ours \\ 
\toprule
Medium-Expert & HalfCheetah & 94.11 & \textbf{96.59} & 87.60 & 93.40 & 95.81\scriptsize{$\pm$ 0.25} & 94.21 \scriptsize{$\pm$ 0.46} & 96.12\scriptsize{$\pm$ 0.18} \\ 
Medium-Expert & Hopper & 113.13 & 113.22 & 98.41 & 111.18 & 115.92\scriptsize{$\pm$ 1.26} & 108.32 \scriptsize{$\pm$ 0.95} & \textbf{118.39\scriptsize{$\pm$ 0.75}} \\ 
Medium-Expert & Walker & 109.90 & 112.21 & 100.52 & 108.71 & 114.87\scriptsize{$\pm$ 0.56} & 111.36 \scriptsize{$\pm$ 0.46}& \textbf{118.51\scriptsize{$\pm$ 0.48}} \\ 
\toprule
Medium & HalfCheetah & 43.90 & \textbf{48.93} & 34.60 & 42.73 & 45.12\scriptsize{$\pm$ 0.34} & 42.28 \scriptsize{$\pm$ 1.18} & 45.51\scriptsize{$\pm$ 0.26} \\ 
Medium & Hopper & 73.84 & 70.44 & 56.98 & 69.42 & 70.45\scriptsize{$\pm$ 0.45} & 72.58 \scriptsize{$\pm$ 0.54} & \textbf{83.24\scriptsize{$\pm$ 0.33}} \\ 
Medium & Walker & 82.05 & 86.91 & 70.95 & 74.70 & 88.71\scriptsize{$\pm$ 0.55} &85.96 \scriptsize{$\pm$ 0.46} & \textbf{88.94\scriptsize{$\pm$ 0.61}} \\ 
\toprule
Medium-Replay & HalfCheetah & 42.27 & 45.84 & 38.81 & 40.31 & \textbf{46.86\scriptsize{$\pm$ 0.33}} & 41.28 \scriptsize{$\pm$ 0.21} & 45.58\scriptsize{$\pm$ 0.36}\\ 
Medium-Replay & Hopper & 90.57 & 98.12 & 78.90 & 88.74 & 96.85\scriptsize{$\pm$ 0.41} &91.32 \scriptsize{$\pm$ 0.66} & \textbf{98.59\scriptsize{$\pm$ 0.26}} \\ 
Medium-Replay & Walker & 76.09 & 91.17 & 65.94 & 68.22 & 92.32\scriptsize{$\pm$ 1.21} & 89.21 \scriptsize{$\pm$ 1.42}& \textbf{96.22\scriptsize{$\pm$ 1.06}} \\
\toprule
\multicolumn{2}{c|}{Total Average} & 80.65\scriptsize{$\pm$ 1.34} & 84.83\scriptsize{$\pm$ 1.10} & 70.28\scriptsize{$\pm$ 1.20} & 77.49\scriptsize{$\pm$ 1.45} & 85.21\scriptsize{$\pm$ 1.12} & 82.84\scriptsize{$\pm$ 0.70} & \textbf{87.90\scriptsize{$\pm$ 1.06}}\\
\toprule
\specialrule{0em}{1.0pt}{1.0pt}
\toprule
\end{tabular}
}
\end{center}
\label{d4rl}
\end{table*}

\begin{figure}
    \centering
    \includegraphics[width=0.85\columnwidth]{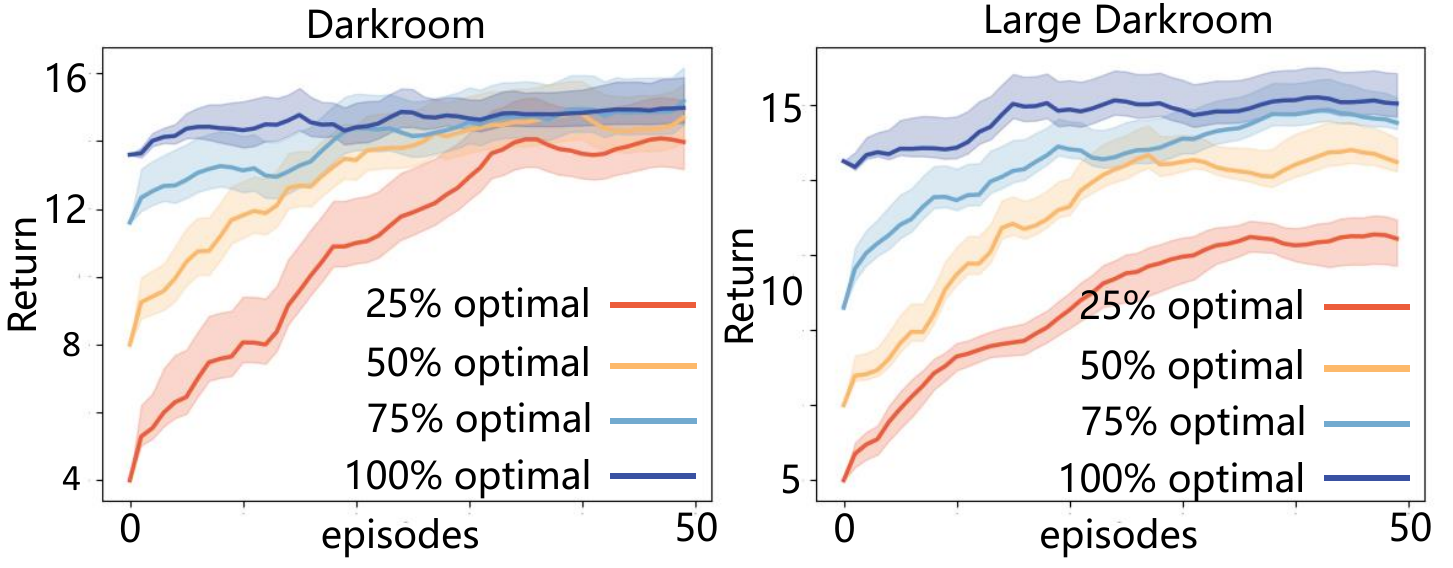}
    \caption{Results for IDT conditioned on partial demonstrations. IDT can accelerate self-improvement through the Review Decisions module to encode external data prompts. 
    }
    \label{partial demonstrations}
\end{figure}

\subsection{Grid World Results}
To evaluate IDT’s self-improvement capabilities in unseen tasks, we compared recent in-context RL methods in the Grid World environments. The agent is required to solve an unseen task by interacting with the environments for 50 episodes without online model updates. As shown in Figure~\ref{grid world}, IDT achieves state-of-the-art performance in a wide range of tasks.

In large variant tasks, IDT significantly surpasses the baselines in both efficiency and performance. However, neither AT nor AD showed obvious self-improvement trends, especially in Large Darkroom Hard. This is because the Large Darkroom Hard is a task with sparse rewards, which makes it difficult for AT and AD to capture the goal position in long sequences. In contrast, IDT explores tasks in a high-level trial-and-error manner, making receiving positive feedback on rewards easier. Overall, IDT demonstrated that a high-level trial-and-error manner is feasible rather than limited to the smallest actions.


\subsection{D4RL Results}
In addition to short-horizon tasks specific to in-context RL methods, we also test the performance of IDT on the D4RL dataset, which is commonly used in conventional RL methods. Based on \citet{13}, the results on D4RL are normalized so that 100 denotes an expert policy. Baseline numbers are reported by the AT paper and from the D4RL paper.
As shown in Table~\ref{d4rl}, IDT outperforms baselines in a majority of the tasks and is competitive with the state-of-the-art in the remaining tasks.

In the TD learning and imitation learning categories, TD3+BC is generally the most remarkable algorithm. Compared with them, the superior performance of IDT demonstrates the advantages of using high-level trial-and-error.

\subsection{Case Study on the Reviewing Decisions Module}
A notable ability of transformer-based policy is to address tasks by providing demonstration prompts. Although in-context RL can improve itself without relying on demonstrations, external prompts can speed up the process.
Therefore, we want to investigate whether IDT can benefit from this setting. 
To answer this question, we design a $\epsilon$-greedy policy to collect external data in Darkroom and Large Darkroom tasks, ranging from nearly random to optimal.

As shown in Figure~\ref{partial demonstrations}, IDT improves each policy in context until it is near-optimal. Notably, the more optimal the input policy, the faster IDT improves it until it is optimal. Despite high-level decisions that cannot be directly observed from the external demonstrations, IDT can extract experts' intentions through the Reviewing Decisions module. In particular, we also perform parameter sensitivity analyses on high-level decision frequency ($c$) and context size ($n$ episodes), as shown in Appendix~\ref{additional experiments}.

\section{Conclusion}

In this work, we propose an efficient in-context RL method IDT that treats RL tasks as an across-episodic sequence problem and can improve itself at test time. The idea of human multi-level decision-making inspires IDT and introduces high-level decisions into the sequence prediction process. Unlike current in-context RL methods limited to short-horizon tasks, IDT is also good at standard RL benchmarks, which typically have longer task horizons. On the Grid World and D4RL benchmarks, we show that IDT can outperform baselines in both efficiency and performance.

\section*{Impact Statement}
In terms of the potential broader impact, our work provides a new idea of incorporating high-level decisions in the design of context, which may promote the development of the in-context RL community.
Besides, we do not see any negative ethical and societal impacts of our work while using our method in practice.

\section*{Acknowledgments}
This work was supported by the National Key R\&D Program of China under Grant No. 2021ZD0112500; the National Natural Science Foundation of China under Grant Nos. U22A2098, U19A2065, 62172185, 61976102, 62206105 and 62202200; the International Cooperation Project of Jilin Province under Grant Nos.20220402009GH, U2341229.

\bibliography{reference}
\bibliographystyle{icml2024}
\newpage
\appendix
\onecolumn
\large
\begin{center}
   \emph{Appendix of paper ``In-Context Decision Transformer: Reinforcement Learning via Hierarchical Chain-of-Thought"} 
\end{center} 
\normalsize

\section{Pseudocode of In-context Decision Transformer}
\label{appendix A}

\begin{algorithm}[h!]
\caption{In-context Decision Transformer.}
\label{IDT algorithm}
\begin{algorithmic}[1]
\STATE \textbf{Input:} A dataset of Trajectories, Max Iterations $M$ as training phase, Max episodes $m$ at testing phase, A number of trajectories $n$ in hierarchical chain of experience, A number of steps of low-level actions $c$ for one high-level decision
\STATE \textbf{Output:} The generated low-level actions
\STATE // \textbf{Training}
\FOR{$i=1$ $\textbf{to}$ $M$}
    \STATE Randomly sample $n$ episodes from dataset $s=(\tau^{1},\tau^{2},\dots,\tau^{n})$
    \STATE Sort $n$ episodes ascending according to their returns $\sum_{t=0}^T{r_t^1}\leq\sum_{t=0}^T{r_t^2}\leq\dots\leq\sum_{t=0}^T{r_t^n}$
    \STATE Compute returns-to-go $\hat{R}_t=\hat{R}_0-\sum_{j=0}^tr_j$ for all steps for each episode, where $\hat{R}_0=\sum_{t=0}^T{r_t^n}$
    \STATE The Reviewing Decisions module encodes a high-level decisions $\textbf{z}$ every $c$ steps
    \STATE Concatenate $n$ episodes as a high-level sequence $s_h=(\tau_h^{1},\tau_h^{2},\dots,\tau_h^{n})$ based on Equation~\eqref{hcoe}
    \STATE Build a low-level sequence every $c$ steps based on Equation~\eqref{lcoe}
    \STATE The Making Decisions module predicts the next high-level decision tokens, and then the Decision to Go module predicts the next $c$ steps low-level action tokens for each predicted high-level decision token
    \STATE Train the Reviewing Decisions, Making Decisions, and Decision to Go modules based on the loss of predicted low-level actions end-to-end    
\ENDFOR
\STATE // \textbf{Testing}
\FOR{$i=1$ $\textbf{to}$ $m$}
    \STATE Start a new episode $i$ and reset the timestep $t=0$
    \WHILE{$t\leq T$}
        \STATE The Making Decisions model generates next high-level decision token $\textbf{z}_t^i$ based on the across-episodic context $(\tau_h^{i-3},\tau_h^{i-2},\tau_h^{i-1},\dots,\hat{R}_t^i,o_t^i)$, where $\tau_h^{i}$ is expressed as Equation~\eqref{hcoe}
        \FOR{$k=0$ $\textbf{to}$ $c-1$}
            \STATE The Decisions to Go model generates next low-level action $a_{t+k}^i$ based on the previous context $(\textbf{z}_t^i,o_t^i,a_t^i,r_t^i,\dots,\textbf{z}_{t}^i,o_{t+k}^i)$
        \ENDFOR
        \STATE The Review Decisions model encode the executed decision $\textbf{z}_t^i$ from the $c$ steps $(o_t^i,a_t^i,\dots,o_{t+c-1}^i,a_{t+c-1}^i)$
        \STATE Compute the sum of $c$ steps rewards $\hat{r}_t^i$ and the next returns-to-go $\hat{R}_{t+c}^i$
        \STATE Receive the next observation $o_{t+c}^i$
        \STATE Update the across-episodic context $((\tau_h^{i-3},\tau_h^{i-2},\tau_h^{i-1},\dots,\hat{R}_t^i,o_t^i,\textbf{z}_t^i,\hat{r}_t^i,d_t^i,\hat{R}_{t+c}^i,o_{t+c}^i)$
        \STATE Update time step $t = t+c$
    \ENDWHILE
\ENDFOR
\end{algorithmic}
\end{algorithm}

In Algorithm~\ref{IDT algorithm}, we introduce the training and testing process of IDT. At each iteration, we first construct a sequence consisting of high-level decisions, as described in lines 5-9. Importantly, high-level decisions in the dataset are encoded by the Reviewing Decisions module (line 8). In addition, each high-level decision will correspond to a short sequence of $c$ steps low-level actions, as described in line 10. Based on the constructed sequences, the Making Decisions and Decisions to Go modules predict high-level decisions and low-level actions, respectively (line 11). Finally, the low-level actions are evaluated with either cross-entropy loss or mean-squared error, depending on whether the actions are discrete or continuous. The losses from each time step are averaged and updated in all three modules end-to-end, as described in line 12.

During testing, IDT needs to generate low-level actions autoregressively and interact with the environment $m$ episodes. At step $t$ of episode $i$ (line 18), the Making Decisions module first generates a high-level decision token $\textbf{z}_t^i$ conditioned on the across-episodic context $(\tau_h^{i-3},\tau_h^{i-2},\tau_h^{i-1},\dots,\hat{R}_t^i,o_t^i)$, where $\tau_h^{i}$ is expressed as Equation~\eqref{hcoe}. Then, the Decision to Go will generate the following $c$ steps low-level actions $(a_t,\dots,a_{t+c-1})$ autoregressively, as described in lines 19-21. Unlike training, the Reviewing Decisions module encodes the executed decision from the actions generated by the Decision to Go module. Then, it serves as a condition for generating the next high-level decision $\textbf{z}_{t+c}^i$, as described in lines 22-26.

\section{Experimental Details}
\label{appendix B}

Source code is available at \href{https://github.com/SiliHuang-ai/IDT}{here}.

\noindent\textbf{Compute.}
Experiments are carried out on NVIDIA GeForce RTX 3090 GPUs and NVIDIA A10 GPUs.
Besides, the CPU type is Intel(R) Xeon(R) Gold 6230 CPU @ 2.10GHz.
Since our memory is not enough to support AT training in D4rl tasks, we refer to the results of the original paper. In contrast, our method has lower memory requirements because it naturally shortens the across-episodic contexts.

\noindent\textbf{Hyperparameters.}
The default length of across-episodic is four trajectories unless mentioned otherwise. In D4RL and Large Grid World, the Decisions to Go module generates $c=10$ steps low-level actions while the Making Decisions module generates one high-level decision. In conventional Grid World, we set $c=5$ because the task is too short. Except for independent parameters and different input and output dimensions, three modules in IDT follow the same architecture. In summary, Table~\ref{hyper} shows the hyperparameters used in our IDT model.

\begin{table}[t]
\centering
\caption{Hyperparameters of IDT.}
\resizebox{0.8\textwidth}{!}{
\begin{tabular}{@{}l|l|l@{}}
\toprule
 & Hyperparameters & Value \\ \midrule
 & Number of layers & 3 \\
 & Number of attention heads & 3 \\
 & Embedding dimension & 128 \\
 & Activation function & ReLU \\ 
 & $c$ steps controlled by one high-level decision & 10 D4RL and Large Grid World \\
 &  & 5 Grid World \\ \midrule
\multirow{7}{*}{Training}& Batch size & 64 \\
 & Dropout & 0.1 \\
 & Learning rate & 1e-4 \\
 & Learning rate decay & Linear warmup for 1e5 steps \\ 
 & Grad norm clip & 0.25 \\
 & Weight decay & 1e-4 \\
 & Number of trajectories to form across-episodic contexts $n$ & 4 (Large) Dark Key-to-Door \\
 &  & 10 other tasks in Grid World \\
 &  & 4 D4RL \\ \midrule
\multirow{14}{*}{Testing} & Target return for HalfCheetah & 12000 \\
 & Target return for Hopper & 3600 \\
 & Target return for Walker & 5000 \\
 & Target return for Darkroom & 20 \\
 & Target return for Darkroom Hard & 1 \\
 & Target return for Darkroom Dynamic & 20 \\
 & Target return for Darkroom Key-to-Door & 2 \\
 & Target return for Large Darkroom & 15 \\
 & Target return for Large Darkroom Hard & 1 \\
 & Target return for Large Darkroom Dynamic & 15 \\
 & Target return for Large Darkroom Key-to-Door & 2 \\
 & Number of trajectories to form across-episodic contexts $n$ & 4 (Large) Dark Key-to-Door \\
 &  & 10 other tasks in Grid World \\
 &  & 4 D4RL \\ 
 \bottomrule
\end{tabular}
}
\label{hyper}
\end{table}

\section{Additional Experimental Results}
\label{additional experiments}

\begin{figure}
    \centering
    \includegraphics[width=0.8\columnwidth]{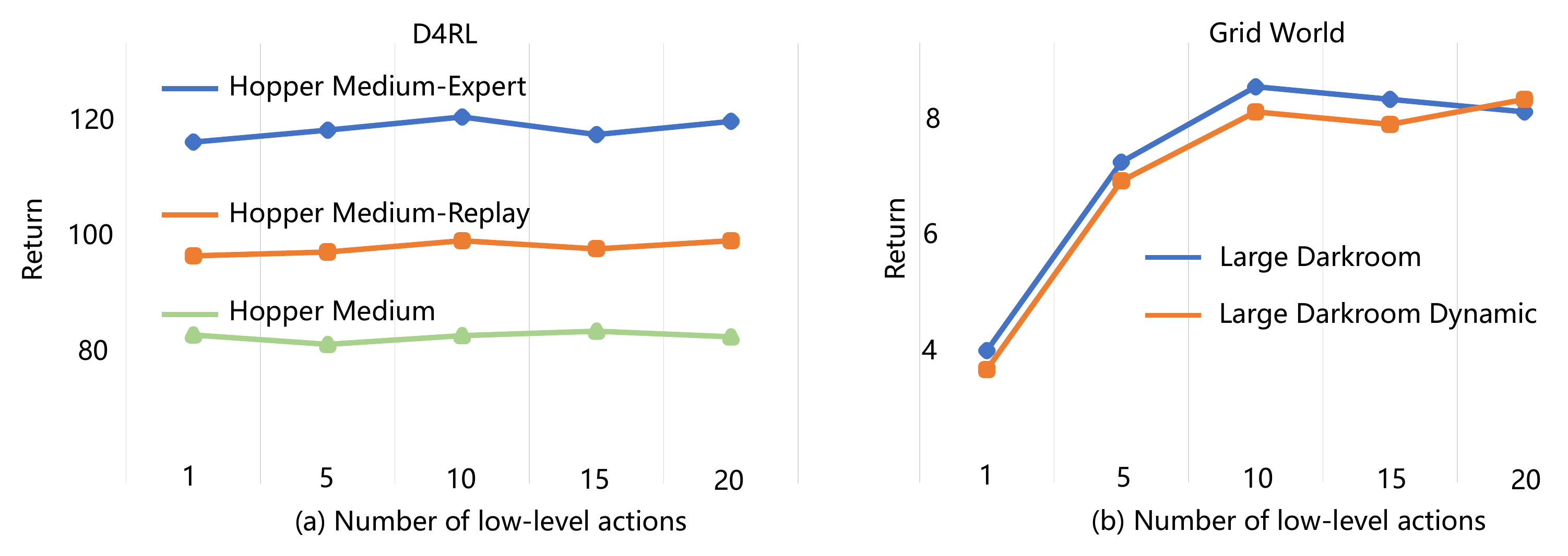}
    \caption{Parameter sensitive analysis of $c$. (a) IDT maintains stable performance to changes in $c$ in dense reward D4RL tasks. (b) As $c$ increases, it becomes easier for the model to receive positive feedback to discover the target location in sparse reward Grid World tasks.}
    \label{horizon}
\end{figure}

\begin{figure}
    \centering
    \includegraphics[width=0.5\columnwidth]{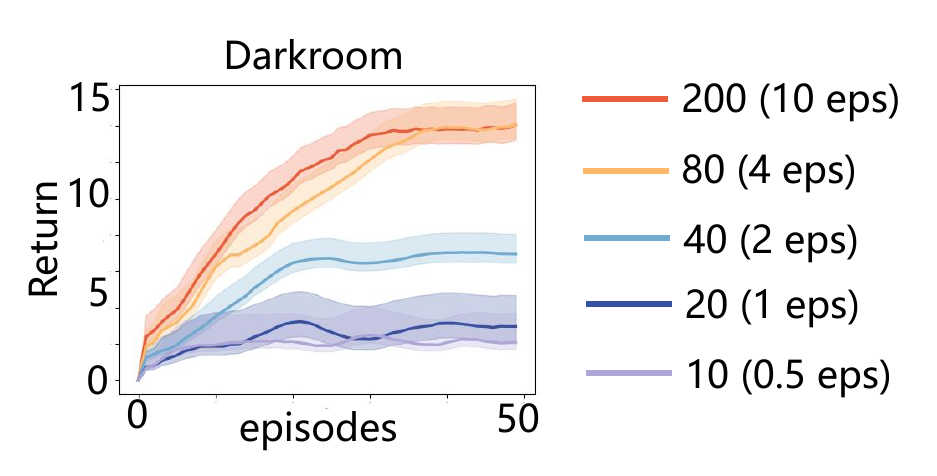}
    \caption{Context size: IDT in Darkroom with different context sizes. IDT emerges with trial-and-error ability once the context size is large enough and across-episodic.}
    \label{Context size}
\end{figure}

\begin{table*}
\caption{Results for training and testing times. We report the training time per 10k gradient updates, the testing time for 50 episodes over Grid World, and 10 episodes over D4RL. As the task length increases, the context length is forced to grow exponentially, resulting in a square increase in computational costs. In contrast, IDT completes trial-and-error on high-level decisions in sizes smaller than one episode length, significantly reducing computational costs.
}
\begin{center}
\resizebox{0.9\textwidth}{!}{
\begin{tabular}{l|l|rrr|rrr}
\toprule
\specialrule{0em}{1.0pt}{1.0pt}
\toprule
\multirow{2}{*}{Context size (step)}&\multirow{2}{*}{Tasks} & \multicolumn{3}{c|}{Training (hour)} & \multicolumn{3}{c}{Testing (minute)} \\ 
\cline{3-8}
 & & AT & AD & Ours & AT & AD & Ours \\ 
 \toprule
\multirow{4}{*}{200} & Darkroom & 0.27 & 0.23 & 0.21 & 0.62 & 0.61 & 0.65 \\ 
& Darkroom Hard & 0.29 & 0.28 & 0.20 & 0.59 & 0.56 & 0.58 \\ 
& Darkroom Dynamic & 0.33 & 0.31 & 0.21 & 0.65 & 0.62 & 0.67 \\ 
& Dark Key-to-Door & 1.12 & 1.01 & 0.44 & 1.89 & 1.50 & 1.52 \\ 
\toprule
\multirow{4}{*}{2000} & Large Darkroom & 5.09 & 4.70 & 2.49 & 67.22 \scriptsize{(13$\times$)} & 45.08 \scriptsize{(9$\times$)} & 5.27 \\ 
& Large Darkroom Hard & 6.48 & 6.69 & 2.93 & 66.81 \scriptsize{(11$\times$)} & 44.96 \scriptsize{(7$\times$)} & 6.09 \\ 
& Large Darkroom Dynamic & 5.71 & 5.84 & 2.73 & 62.06 \scriptsize{(11$\times$)} & 42.12 \scriptsize{(8$\times$)} & 5.51 \\ 
& Large Dark Key-to-Door & 18.87 & 18.23 & 3.06 & 167.07 \scriptsize{(27$\times$)} & 76.79 \scriptsize{(12$\times$)} & 6.18 \\
\toprule
\multirow{3}{*}{4000} & HalfCheetah & 36.18 & 37.10 & 21.90 & 234.20 \scriptsize{(37$\times$)} & 173.11 \scriptsize{(28$\times$)} & 6.29 \\
& Walker & 32.82 & 33.77 & 20.08 & 233.18 \scriptsize{(36$\times$)} & 172.34 \scriptsize{(26$\times$)} & 6.51 \\
& Hopper & 24.08 & 22.23 & 12.99 & 232.82 \scriptsize{(35$\times$)} & 172.92 \scriptsize{(26$\times$)} & 6.56 \\
\toprule
\specialrule{0em}{1.0pt}{1.0pt}
\toprule
\end{tabular}
}
\end{center}
\label{cost}
\end{table*}

\noindent\textbf{Parameter Sensitive Analysis of $c$.}~~
An important insight of IDT is that one high-level decision can guide $c$-step low-level actions. We aim to investigate whether the size of $c$ will affect the performance of IDT. Therefore, we tested $c=1,5,10,15,20$ in D4RL and Grid World, respectively. As shown in Figure~\ref{horizon}, IDT maintains stable performance to changes in $c$ in D4RL tasks. In contrast, larger $c$ achieves better performance in Grid World. This is because the Grid World is designed for tasks with sparse rewards where the agent needs to rely on rewards to reason about the target location. As $c$ increases, it becomes easier for the model to receive positive feedback to discover the target location.

\noindent\textbf{What Context Size is Required for IDT?} Similar to other in-context RL methods, we also test how context sizes are required for IDT emerging with trial-and-error ability. As shown in Figure~\ref{Context size}, multi-episodic contexts of 4 episodes are necessary to learn a near-optimal IDT. When the context size is roughly the length of an episode, IDT begins to emerge with self-improvement. The reason for this is likely that the context is large enough to retrain across-episodic information – e.g., at the start of a new episode, the context will be filled with transitions from most of the previous episode.

\end{document}